\documentclass[letterpaper,twocolumn]{article}
\usepackage{aaai16}
\usepackage{times}
\usepackage{helvet}
\usepackage{courier}

\nocopyright

\usepackage{graphicx}
\usepackage{enumitem}
\usepackage{amsmath}
\usepackage{color}

\usepackage{bigstrut}
\usepackage{array}

\usepackage{natbib}

\renewcommand\cite{\citep}

\newcommand{\oyvind}[1]{{\texttt \small {\color{cyan} [OT: {#1}]}}}

\newcommand{\bful}[1]{\textbf{\underline{#1}}}
\newcommand{\eat}[1]{}

\usepackage[dvipsnames]{xcolor}	

\newcommand{\red}[1]{\textcolor{red}{#1}}

\newcommand{\green}[1]{\textcolor{ForestGreen}{#1}}

\mathchardef\mhyphen="2D
\newenvironment{des}{                 
     \parskip 0cm \begin{list}{}{\parsep 0cm \itemsep 0cm \topsep 0cm}}{
       \end{list}} 

\usepackage{quoting}

\newcommand{\quotebox}[1]{\begin{quote}\fbox{\parbox{0.9\columnwidth}{#1}}\end{quote}}

\newcommand\macaw{\textsc{Macaw}}
\newcommand{\macawurl}{https://github.com/allenai/macaw}

\title{General-Purpose Question-Answering with \macaw{}}

\author{
  Oyvind Tafjord and
  Peter Clark
         \\
         \ \\
Allen Institute for Artificial Intelligence, Seattle, WA, U.S.A.\\
  {\tt \{oyvindt,peterc\}@allenai.org}
  }

\date{\today}

\begin{document}
\maketitle

\begin{abstract}

  Despite the successes of pretrained language models, there are still few high-quality,
  general-purpose QA systems that are freely available. In response, 
  we present \macaw{}, a versatile, generative question-answering (QA) system that we are making available to the community.
  \macaw{} is built on UnifiedQA, itself built on T5, and exhibits strong performance, zero-shot, on a wide variety of topics, 
  including outperforming GPT-3 by over 10\% (absolute) on Challenge300, a suite of 300 challenge questions,
  despite being an order of magnitude smaller (11 billion vs. 175 billion parameters).
  In addition, \macaw{} allows different permutations (``angles'') of its inputs and outputs
  to be used, for example \macaw{} can take a question and produce an answer; or take an answer and
  produce a question; or take an answer and question, and produce multiple-choice options.
  We describe the system, and illustrate a variety of question types where it produces surprisingly
  good answers, well outside the training setup. We also identify question classes where it still appears to struggle,
  offering insights into the limitations of pretrained language models.
  \macaw{} is freely available, and we hope that it proves useful to the community.\footnote{\macaw{} is available at \macawurl}
\end{abstract}

\section{Introduction}

Although modern pretrained language models have proved surprisingly effective at solving datasets,
e.g., \cite{gpt2,Raffel2020ExploringTL,unifiedqa}, there are still few high-quality, general-purpose, off-the-shelf
question-answering (QA) systems freely available. UnifiedQA \cite{unifiedqa} is a powerful QA system, but
mainly trained for span prediction and multiple-choice selection
rather than answer generation.
GPT-3 appears powerful, but is not freely available to the public \cite{gpt3}.
One nearest to our goal is Google's T5-based CBQA (closed-book QA) system \cite{Roberts2020HowMK},
but in our tests of the T5-CBQA model trained on Natural Questions \cite{47761}, it did not perform as well as \macaw{}
(Section~\ref{challenge300-section}).\footnote{There are other T5-CBQA versions alternatively trained on WebQuestions and TriviaQA that we did not evaluate, although NaturalQuestions is arguably the most general and varied of the alternate training sets. See https://github.com/google-research/google-research/tree/master/t5\_closed\_book\_qa}
Similarly, the new Jurassic-1 model from AI21 Labs\footnote{https://www.ai21.com/blog/announcing-ai21-studio-and-jurassic-1}
appears powerful, but again did not perform as well in our QA tests (Section~\ref{challenge300-section}).
In response, we are releasing \macaw{}, a
general-purpose, T5-based QA system that we have found to be to be surprisingly strong, zero-shot.
In this paper, we describe how \macaw{} was trained, illustrate the variety of capabilities it has,
and also identify various question classes it still struggles with.
We hope \macaw{} proves useful for the community.

\begin{figure}[t]
\centerline{
 \fbox{%
   \parbox{\columnwidth}{
     {\small
Q: How would you make a house conduct electricity? \\
\vspace{1mm}
$\rightarrow$ A: paint it with a metal paint \\
Q: What is an implication of a cat being an animal? \\
\vspace{1mm}
$\rightarrow$ A: it has feelings \\
Q: Could a newborn look over the top of a fully grown horseradish plant? \\
\vspace{1mm}
$\rightarrow$ A: no, because the plant is so tall \\
Q: Why couldn't Aristotle have used a cellphone? \\
\vspace{1mm}
$\rightarrow$ A: he was born before cellphones \\
Q: A young boy was rushed to the hospital emergency room, but the ER doctor saw the boy and refused to operate. "This boy is my son," the doctor said. But the doctor wasn't the boy's father. How could this be? \\
{\it [Intended answer to the riddle: The doctor was the boy's mom]} \\
$\rightarrow$ A: the doctor mistook the boy for his own son {\it [Reasonable alternative answer!]}
}}}}
\caption{A few examples of the more impressive answers given by \macaw{}.
  More examples (both good and bad) are given in Section~\ref{analysis}.}
\label{examples}
\end{figure}

\eat{
\begin{quote}
\fbox{\parbox{0.9\columnwidth}{
Q: A young boy was rushed to the hospital emergency room, but the ER doctor saw the boy and refused to operate. "This boy is my son," the doctor said. But the doctor wasn't the boy's father. How could this be? \\
{\it Intended answer to the riddle: The doctor was the boy's mom} \\
$\rightarrow$ \macaw{}'s Answer: he mistook the boy for his own son}}
\ \\
\fbox{\parbox{0.9\columnwidth}{
Q: A truck driver is going down a one-way street the wrong way and passes at least ten cops. Why is he not caught? \\
{\it Intended answer to the riddle: Because he was not driving! He's walking on the sidewalk.} \\
$\rightarrow$ A: they are not looking}}
\end{quote}
Although \macaw{} does not give the conventional answers to the riddles, both its answers are
completely reasonable. This is particularly remarkable given the complexity of the riddles,
and raises interesting questions about the extent to which the model understands the
scenes the riddles describe.
}

\macaw{} has three interesting features. First, it often produces high-quality answers
to questions far outside the domain it was trained on, sometimes surprisingly so. Several examples are shown in Figure~\ref{examples}, and
we show numerous other examples later in this paper (Section~\ref{impressive}). However, it can also make mistakes.
We also give a selection of these, and attempt to characterize where
its weaknesses are (Section~\ref{unimpressive}).

Second, \macaw{} allows different permutations (``angles'') of inputs and outputs to
be used. For example, we can give it a question and get an answer; or give it an answer
and get a question; or give it a question and answer and get a set of multiple-choice (MC)
options for that question. This multi-angle QA capability\footnote{
  Hence then model's name \macaw{} (``Multi-angle c(q)uestion-answering'').} allows
versatility in the way \macaw{} can be used, include recursively using outputs
as new inputs to the system. While other researchers have explored permuting
inputs and outputs to some degree, e.g., \cite{Hase2020LeakageAdjustedSC},
\macaw{} has such capabilities built into its machinery.

Finally, \macaw{} also generates explanations as an optional output (or even input) element.
Although \macaw{}'s explanations are of typically of lower quality than its answers,
and are not full chains of reasoning, the fact it can generate plausible
explanations at all is an unusual feature.

We first describe multi-angle training and how \macaw{} was trained. 
We then report quantitative experiments with \macaw{}, including comparing its zero-shot
performance with several other large language models on Challenge300, a suite of
300 challenge questions designed to push various limits of question-answering behavior.
We find \macaw{} outperforms other large-scale models by over 10\% (absolute) on this dataset,
including GPT-3 despite being an order-of-magnitude smaller (11B \macaw{} vs. 175B GPT-3).
We then give a qualitative analysis of \macaw{}'s behavior, identifying classes of problems
where it succeeds and also where it fails.
Finally we reflect on its behavior and offer \macaw{} to the community. \macaw{} is available at~\macawurl.

\section{Multi-Angle Question-Answering}

\subsection{Slots, Values and Angles}

We take advantage of the flexible nature of text-to-text transformers like T5 \cite{Raffel2020ExploringTL} to train models across multiple ``angles'' for each dataset. Each example in the dataset is considered as a set of slots $S_i$ and corresponding values $V_i$.
An angle $A_i = \{S_{s_i}\} \rightarrow \{S_{t_i}\}$ then corresponds to a specific set of source slots $S_{s_i}$ and a set of target slots $S_{t_i}$, and the associated task is to predict the values of the target slots given the source values.

For instance, in a multiple-choice QA dataset, like ARC \cite{Clark2018ThinkYH} or RACE \cite{race}, the slots might be Q (question), M (MC options), A (correct answer), C (context). The usual task is represented by the {\it primary angle} QMC$\rightarrow$A (given question, MC choices and context, what is the correct answer?). Other angles might include QC$\rightarrow$A (answer the question without seeing the MC choices), QAC$\rightarrow$M (generate plausible MC choices), AC$\rightarrow$QM (given answer and context, generate a question and answer options). See Figure~\ref{angles} for more examples.

The semantics of slots are defined by what \macaw{} saw during training (Section~\ref{training-macaw}).
During training, the context (C) contains either a passage or retrieved text relevant to the question, and the explanation (E) consists of a few (typically two or three) general sentences relevant to the answer (but not a formal chain of reasoning). Examples are given in Figure~\ref{angles} (upper box) and Section~\ref{macaw-examples}.

\begin{figure}[t]
  \quotebox{
    \begin{des}
    \item[{\bf \underline{Slot:}}] \underline{Example value} 
\item[{\bf C (context, e.g., from IR):}] Roller skating is a popular hobby these days. Roller skates have four wheels....
\item[{\bf Q (question):}] Which surface is best for roller-skating?
\item[{\bf M (multiple-choice (MC) options):}] (A) gravel (B) blacktop (C) sand
\item[{\bf A (answer):}] blacktop
\item[{\bf E (explanation):}] A wheeled vehicle requires smooth surfaces.
\eat{\item[{\bf X (special: output probabilities for each option):}] (A) gravel (B) blacktop (C) sand}
\end{des}}
 \quotebox{
    \begin{des}
    \item[{\bf \underline{Angle:}}] \underline{Description} 
\item[{\bf QMC$\rightarrow$AE:}] Generate answer and explanation given question, choices and context (primary angle).
\item[{\bf QM$\rightarrow$AE:}] Same, but in absence of retrieved context
\item[{\bf QMC$\rightarrow$A:}] Only generate answer
\item[{\bf QC$\rightarrow$A:}] Generate answer without access to MC options
\item[{\bf QMEC$\rightarrow$A:}] Also include explanation in input
\item[{\bf QAC$\rightarrow$M:}] Generate plausible MC options given question, answer and context
\item[{\bf AC$\rightarrow$QM:}] Generate plausible question and MC options, given answer and context
\end{des}}
\caption{The different slots (input/output elements) and sample angles supported by \macaw{}. \label{angles}}
\end{figure}

\eat{
\begin{figure}[t]
  \quotebox{
    \begin{des}
    \item[{\bf \underline{Slot:}}] \underline{Example value} 
\item[{\bf C (context, e.g., from IR):}] When the submarine descends deep into the ocean, you can hear the creaks and groans of the submarine being crushed by the water pressure....
\item[{\bf Q (question):}] As submarines descend, scientists observe that there is an increase in the
\item[{\bf M (multiple-choice (MC) options):}] (A) amount of light (B) water temperature (C) water pressure (D) types of ocean organisms
\item[{\bf A (answer):}] water pressure
\item[{\bf E (explanation):}] As an object descends into water, the pressure on that object will increase
\end{des}}
\caption{The different slots and sample angles for a question from ARC. \label{angles}}
\end{figure}
}

For each dataset we collect a set of angles which would be considered reasonable tasks to perform. E.g., in the RACE dataset the context is usually critical to answer a situated question (e.g., ``What does the doctor think of Heelys?'') so we do not consider the QM$\rightarrow$A angle without the context, while this angle is appropriate for ARC where the context is just a potentially helpful retrieved text passage.

\subsection{Text Encoding of Angles \label{text-encoding}}

We employ a simple text format to encode an angle input/output pair $\{S_{s_1}, S_{s_2}, ...\} \rightarrow \{S_{t_1}, S_{t_2}, ...\}$:

\vspace{2mm}
{\small
\noindent
{\bf INPUT:}  \texttt{"\$$S_{t_1}$\$ ; \$$S_{t_2}$\$ ; ... \$$S_{s_1}$\$ = $V_{s_1}$ ; \$$S_{s_2}$\$ = $V_{s_2}$ ; ..."}\\
{\bf OUTPUT:} \texttt{"\$$S_{t_1}$\$ = $V_{t_1}$ ; \$$S_{t_2}$\$ = $V_{t_2}$ ; ..."}
}

\vspace*{2mm}
\noindent
In other words, in the INPUT field the desired output slots $S_{t_i}$ are listed without
any associated value, and the input slots $S_{s_i}$ are listed with
their corresponding input values.
For instance, to provide the ``question'' and ``mcoptions'' (multiple-choice
options) as inputs, and request the ``answer'' and ``explanation'' slots
in the output, the INPUT format might look as below, resulting in
the corresponding OUTPUT from \macaw{}:

\vspace*{2mm}
\noindent
{\small
{\bf INPUT:}  \texttt{"\$answer\$ ; \$explanation\$ ; \$question\$ = Which surface is best for rollerskating? ; \$mcoptions\$ = (A) gravel (B) sand (C) blacktop"}\\
{\bf OUTPUT:}  \texttt{"\$answer\$ = blacktop ; \$explanation\$ = A wheeled vehicle requires smooth surfaces."}
}

\subsection{Ordering of Slots within an Angle}

We can either treat the input slots as an unordered set or in a certain fixed order. Given the nature of the transformer encoder, it is not expected that the input order has great significance. In practice we scramble the order of the input and output slots during training, except putting the "context" slot at the end as it tends to be the one that might run over the token limit (512 subword tokens in the case of T5). 

If there are multiple output slots, such as producing both answer and explanation, their ordering might carry more significance due to the left-to-right nature of decoding. E.g., first producing explanation followed by answer, is technically generating the answer conditioned on the already generated explanation. Again, for simplicity and practicality, for \macaw{} we train (and evaluate) with randomly scrambled orders of output slots.

\subsection{Sampling of Angles during Training}

We describe the precise training of \macaw{} shortly. During training, we sample the possible angles across the training set rather than considering every angle for every training instance. The training recipe includes the following:

\begin{itemize}
  \item Each angle can have a heuristic scaling factor for how often it is sampled relative to others (used as weak bias for which angles are more meaningful).
  \item We iterate through the training instances  multiple times (especially if there are many angles and not that many training instances)
  \item If a sampled angle does not exist for a training instance (e.g., the explanation value is only available for a subset of instances), the angle is resampled. This allows handling of heterogenous datasets where slots are partially available.
\end{itemize}

For evaluation we generate all angles for every instance (with random scrambling of the slot orders if that was the chosen mode during training, as was done for the \macaw{} model).

\subsection{Decoding and Evaluation}

\macaw{}'s default decoding is done with greedy decoding, optionally with a small beam search, which is appropriate for well-defined slot values like answers. For more open-ended slot values, like question generation, \macaw{} also supports sampling (e.g., nucleus sampling \cite{Holtzman2020TheCC}), allowing alternate outputs to be generated.

When the full output string has been generated, e.g., Section~\ref{text-encoding}, it is straightforward to parse it with a regular expression pattern to extract the produced slots and values. These can then be evaluated according the their usual metrics. If an expected slot is missing, it is counted as a failure, but in practice this almost never happens.

An alternative decoding approach, most suitable for multiple-choice answers, is to forcibly generate each answer choice, and the product of probabilities of the output tokens then provide a confidence score for each answer choice.

\subsection{Examples of Different Angles \label{macaw-examples}}

Some examples of using \macaw{} with different angles are shown below, where the
right arrow (``$\rightarrow$'') indicates \macaw{}'s generated output, given the inputs.
Note that some of these angles were not seen during training (in fact, only the first two are
seen during training), yet \macaw{} is still able to reliably handle them. 
(The full list of angles seen in training is described shortly in Section~\ref{training} and Table~\ref{macaw-datasets}).
\begin{description}
\item[\bful{ Q$\rightarrow$A}:] Given a question, produce an answer.
\quotebox{
Q: What gas do producers produce that consumers need to survive? \\
$\rightarrow$ A: oxygen}
\item[\bful{ Q$\rightarrow$AE}:] Given a question, produce an answer + explanation.
\quotebox{
Q: Which force pulls objects to the ground? \\
$\rightarrow$ A: gravity \\
$\rightarrow$ E: Gravitational force causes objects that have mass to be pulled down on a planet.}
\item[\bful{ A$\rightarrow$QE}:] Given an answer, produce a question + explanation.
\quotebox{
A: elephant \\
$\rightarrow$ Q: Which animal has the largest ears? \\
$\rightarrow$ E: The ears of an elephant are the largest.}
\item[\bful{ A$\rightarrow$QM}:] Given an answer, produce a question + multiple-choice options.
\quotebox{
A: car battery \\
$\rightarrow$ Q: Which of these items is necessary for a car to start? \\
$\rightarrow$ M: (A) car battery (B) windshield wiper blade (C) car radio (D) car radio antenna}
\item[\bful{ E$\rightarrow$QA}:] Given an explanation, generate a question + answer.
\quotebox{
E: The leaves of a plant convert sunlight into food for the plant. \\
$\rightarrow$ Q: How do plants get energy? \\
$\rightarrow$ A: from the sun}
\end{description}

\subsection{Output Confidences}

\macaw{} can also output the generation scores for a set of candidate answers,
conventionally invoked with a ``X'' (eXplicit outputs) slot, e.g.,:
\quotebox{
Q: What is the largest animal in the world? \\
X: (A) mouse (B) whale (C) elephant \\
$\rightarrow$ A: whale (0.007), elephant (0.005), mouse (1.4e-8)}
Note the confidences do not add to one, as {\it other} answers (e.g., ``blue whale'') are
possible but are not listed. To further condition the naturally generated answers, the question can be formulated as multiple-choice using the ``M'' slot as well:
\quotebox{
Q: What is the largest animal in the world? \\
M: (A) mouse (B) whale (C) elephant \\
X: (A) mouse (B) whale (C) elephant \\
$\rightarrow$ A: whale (0.999), elephant (3.9e-5), mouse (2.4e-11)}
In this case the confidences, which are the product of the internal output token probabilities, do tend to add up to one as the model is strongly biased towards picking one of the answers in from the ``M'' slot.

\section{Training \macaw{} \label{training-macaw}}

\macaw{} is built on top of the text-to-text pretrained T5 transformer \cite{Raffel2020ExploringTL}, by first
training a multi-angle version version of UnifiedQA \cite{Khashabi2020UnifiedQACF}, followed by further fine-tuning on science questions with explanations, using the ARC \ and ARC-DA datasets along with explanations from WorldTree \cite{Jansen2018WorldTreeAC}.

\subsection{Multi-Angle UnifiedQA \label{multi-angle-uqa}}

\begin{table}
\centering
{\small
\setlength{\tabcolsep}{3pt}	
\begin{tabular}{p{80pt}p{140pt}} \hline
{\bf Datasets}	& {\bf Angles}  \\\hline
BoolQ, NarrativeQA, SQuAD 2.0 &  QC$\rightarrow$A, AC$\rightarrow$Q \\\hline
ARC, OBQA & QMC$\rightarrow$A, QC$\rightarrow$A, QM$\rightarrow$A, \newline QAC$\rightarrow$M, MAC$\rightarrow$Q, AC$\rightarrow$QM \\\hline
RACE, MCTest & QMC$\rightarrow$A, QC$\rightarrow$A, QAC$\rightarrow$M, \newline MAC$\rightarrow$Q \\\hline
\end{tabular}
}
\caption{Datasets and angles used in training of multi-angle UnifiedQA (the slots are Q=Question, C=Context, M=MC options, A=Answer). \label{unifiedqa-datasets}}
\end{table}

The multi-angle version of UnifiedQA was trained on the 7 core datasets with associated angles listed in Table~\ref{unifiedqa-datasets}. The 11B model was finetuned for 120k steps starting from T5-11B with batch size of 8 and the Adafactor optimizer. These datasets vary greatly in size (from 1.5k to 130k training instances), following UnifiedQA we sample equally from the 7 datasets. For direct comparison we also trained a similar single-angle version using the same setup. 

For the ARC and OBQA datasets, the context (``C'') consists of 10
sentences retrieved from a general text corpus based on the question text plus each of multiple-choice options (ranked by IR score, but always keeping the top result for each option).\footnote{
For this we use the Aristo Corpus, a Web-crawled corpus containing 280GB of general and
science-related sentences augmented with $\approx$80k additional science textbook sentences \cite{Clark2016CombiningRS}.} 

The performance of these models on the primary angle is very similar to the original UnifiedQA model. Table \ref{unifiedqa-eval} shows a comparison between the scores of the single-angle and multi-angle models on the development sets, showing the multi-angle is generally not much behind the single-angle variant, while providing more general functionality through the alternate angles. 

We train multi-angle UnifiedQA in three sizes based on T5-11B, T5-3B, and T5-large. As seen in Table~\ref{unifiedqa-eval}, for some of the datasets there is a significant drop in evaluation scores for smaller sizes, but the scores are still high in general.

\begin{table}
\centering
{\small
\begin{tabular}{l|c|ccccc}
\hline
{\bf Model$\rightarrow$} & {\bf Single-Angle} & \multicolumn{3}{|c}{\bf Multi-Angle UnifiedQA} \\
{\bf Dataset$\downarrow$} & {\bf 11B} & {\bf 11B} & {\bf 3B} & {\bf large (770M)}\\
\hline
BoolQ & 90.8 & 90.3 & 89.1 & 85.4\\
NarrativeQA & 66.5 & 66.8 & 65.4 & 62.8\\
SQuAD 2.0 & 91.1 & 90.3 & 89.4 & 86.8\\
\hline
ARC & 88.6 & 87.0 & 81.9 & 72.2\\
MCTest & 96.6 & 95.9 & 94.4 & 90.9\\
OBQA & 87.4 & 88.4 & 81.8 & 71.4\\
RACE & 88.0 & 87.7 & 84.4 & 79.2\\
\hline
\end{tabular}
}
\caption{Model performance (averaged over UnifiedQA datasets (dev partition), measuring accuracy except for SQuAD 2.0 (token F1) and NarrativeQA (ROUGE-L)). Multi-angle UnifiedQA retains performance compared with single-angle (for same size models, columns 1 and 2), while adding multi-angle capabilities.
  Evaluation is on the primary angle (QC$\rightarrow$A for the 3 first datasets, QMC$\rightarrow$A for the other 4).
  ARC includes both the Easy and Challenge categories.
\label{unifiedqa-eval}}
\end{table}

\subsection{\macaw{} \label{training}}

For the final \macaw{} model, we further fine-tune multi-angle UnifiedQA on the ARC dataset as well as the ARC-DA dataset, a dataset of Direct Answer (``open response'', ``freeform'')
science questions \cite{arc-da} (with 1250 questions in the training set). 

For each question we add an input context (``C'') based on retrieval from a text corpus as described in the previous section (for ARC-DA the retrieval is based only on question text as there are no answer options available).

We also add an explanation (``E'') to each question using data from the WorldTree V2 explanation bank \cite{Jansen2018WorldTreeAC}. WorldTree contains explanations for a subset of the questions in ARC and ARC-DA (the fraction of questions covered is about 65\% for ARC and 50\% for ARC-DA). We construct a short explanation paragraph by randomly shuffling the sentences marked as ``CENTRAL'' (in the few cases with more than 5 such sentences, we sample 5 of them). 

\begin{table}
\centering
{\small
\setlength{\tabcolsep}{3pt}	
\begin{tabular}{p{40pt}p{180pt}} \hline
{\bf Dataset}	& {\bf Angles}  \\\hline
ARC &  QMC$\rightarrow$AE, AQC$\rightarrow$M, 
CQME$\rightarrow$A, QME$\rightarrow$A, \newline
QE$\rightarrow$A,
QMC$\rightarrow$A, QC$\rightarrow$AE, QM$\rightarrow$AE,\newline QMAC$\rightarrow$E, QMA$\rightarrow$E\\\hline
ARC-DA & QC$\rightarrow$AE, Q$\rightarrow$AE, QC$\rightarrow$A, Q$\rightarrow$A, CQE$\rightarrow$A, \newline
QE$\rightarrow$A, AE$\rightarrow$Q, AC$\rightarrow$Q, QA$\rightarrow$E, AQC$\rightarrow$E \\\hline
\end{tabular}
}
\caption{Datasets and angles used in training of \macaw{} (with slots as in Table~\ref{unifiedqa-datasets} plus E=Explanation). \label{macaw-datasets}}
\end{table}

With five available input/output slots there is a plethora of possible angles to train on. We select a subset that seem the most interesting, as listed in Table~\ref{macaw-datasets}, and use these for fine-tuning for 6k further steps.

\begin{figure*}
\centering
\includegraphics[width=1\textwidth]{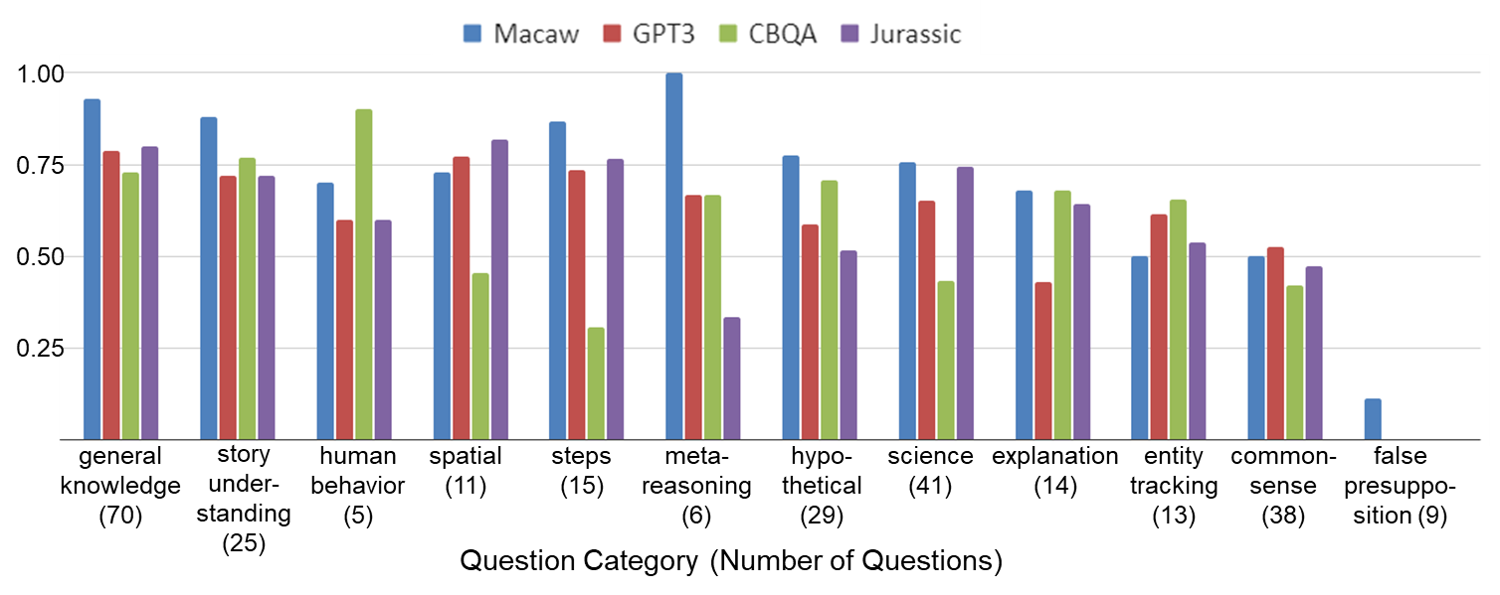}	   
\vspace{-7mm}
     \caption{Average score of the four different models on different categories of questions (ignoring
       categories with less than five questions). The numbers in parentheses denotes the number of questions in each category.
       Categories are ordered by average-of-averages (highest to lowest), i.e.,, the models together perform best on general knowledge and worst on false presuppositions.
       The tabular version of this data is in the Appendix.
       \label{challenge300-results-figure}}
\end{figure*}

\section{Quantitative Performance of \macaw{}}

\begin{table}
\centering
{\small
\setlength{\tabcolsep}{3pt}	
\begin{tabular}{lcccccc}
\hline
{\bf Angle$\rightarrow$} & QMC & QMC &  QMEC & QM &  QM &  QME\\
{\bf Dataset+Model$\downarrow$} & $\rightarrow${\bf A}E & 
$\rightarrow${\bf A} &  $\rightarrow${\bf A} &
$\rightarrow${\bf A}E & 
$\rightarrow${\bf A} &  $\rightarrow${\bf A}\\
\hline
{\bf ARC-Easy:}\\
\hspace*{1mm}\macaw{}  (11B) & 90.9 & 91.2 & 94.0 & 85.1 & 84.9 & 91.4\\
\hspace*{1mm}\macaw{}-3B & 87.5 & 87.9 & 91.6 & 77.7 & 76.7 & 85.3 \\
\hspace*{1mm}\macaw{}-large & 82.5 & 82.5 & 86.1 & 66.3 & 63.9 & 79.8 \\
\hline
{\bf ARC (Challenge):}\\
\hspace*{1mm}\macaw{}  (11B) & 76.9 & 76.9 & 86.3 & 74.6 & 74.6 & 86.6\\
\hspace*{1mm}\macaw{}-3B & 69.6 & 68.2 & 80.9 & 66.2 & 67.9 & 77.9 \\
\hspace*{1mm}\macaw{}-large & 53.9 & 57.2 & 66.9 & 48.2 & 50.5 & 67.2 \\
\hline
\end{tabular}
}
\caption{Scores for the answer output slot {\bf A} on ARC (Easy and Challenge) multiple-choice development sets, for six different angles.
\label{arc-mc-eval}}
\end{table}

\subsection{The ARC dataset \label{arc}}

While this paper mainly focuses on an analysis of \macaw{}'s capabilities and limitations,
we note that the "answer-focused" variant \macaw{}-answer-11B is at the time of publication at the top of the leaderboards
for the datasets ARC (with a score of 81.4\%),\footnote{https://leaderboard.allenai.org/arc/submissions/public} ARC-Easy (92.7\%),\footnote{https://leaderboard.allenai.org/arc\_easy/submissions/public} and 
ARC-DA (81\%).\footnote{https://leaderboard.allenai.org/genie-arcda/submissions/public} This variant was trained without the explanation slot and with a focus on the answer-output angles. This model is also available in our software release.\footnote{https://github.com/allenai/macaw} 

\eat{
CommonsenseQA\footnote{https://www.tau-nlp.org/csqa-leaderboard} \cite{Talmor2019CommonsenseQAAQ},
and OpenbookQA\footnote{https://leaderboard.allenai.org/open\_book\_qa/submissions/public} \cite{OpenBookQA2018}.
}

To get a sense of the variation with model size, Table~\ref{arc-mc-eval} 
gives scores on the ARC development set for the smaller \macaw{}-3B and \macaw{}-large in addition to the default \macaw{} (11B). There is a clear performance drop for smaller sizes, but the smaller models still provide good performance and might be more practical for deployment and further research iterations.

In Table~\ref{arc-mc-eval}, we observe that if the answer explanation is included
in the {\it input} angle (result columns 3 and 6), the answer accuracy significantly improves.
This is perhaps not surprising as the explanation typically strongly hints at (or even includes) the right answer (the effect is larger than indicated in the tables, since only a subset of questions actually have explanations in the dataset). 

One could hypothesize that feeding the model's own explanation back as input could also help (first run QC$\rightarrow$AE, then use the E output as input to QEC$\rightarrow$A), but from small-scale tests this generally only had a minor effect on the score, while tending to make the precision-recall curves look worse (presumably because originally uncertain, incorrect answers, will now get reinforced through the explanation to higher confidence).

\eat{
\oyvind{Add PR curve for 11B model on ARC Challenge running in X mode, with and without roundtripping explanations?}

In Table~\ref{arc-mc-eval-other-angles} we show examples of evaluation of output slots other than answers, using ROUGE-L as a rough evaluation metric. \oyvind{Need better data for this, possibly some other interesting angles, would be good to have some examples of evaluating these angles, but could leave out}
}

\begin{table*}
\centering
{\small
\setlength{\tabcolsep}{3pt}	
\begin{tabular}{llp{13cm}} \hline 
  {\bf Category}& {\bf \# Qns} & {\bf Description + Example} \\ \hline
  commonsense & 38 & Obvious (to a person) facts about the world \\
  & & {\it If I put some cheese in the fridge, will it melt?} \\
comparison & 2 & Relation between two entities  \\
 & & {\it How do pandas and parrots differ? } \\
entity substitution & 4 & Find a suitable replacement entity for a task \\
 & & {\it How would you bang in tent pegs without a hammer? } \\
entity tracking & 13 & Tracking entity states over time \\
 & & {\it My house is red. I painted my house white. What color is my house now? } \\
estimation & 4 &  Fermi-style numeric estimation problems \cite{fermi} \\
 & & {\it How many banknotes can you fit in a school bus? } \\
example generation & 2 &  Create an illustration of a general phenomenon \\
 & & {\it If you let go of an object, then gravity pulls it to the ground. What is an example of this phenomenon? } \\
explanation & 14 &  ``Why...?'' questions \\
 & & {\it Why do houses have roofs? } \\
false presupposition & 9 &  Trick questions that presuppose something that is not true \cite{Kim2021WhichLI} \\
 & & {\it What year did Tom Hanks land on the moon? } \\
general knowledge & 70 & General facts about the world \\
 & & {\it What is shiplap? } \\
generation & 1 & Production of prose \\
 & & {\it Tell me a story about a car accident. } \\
history & 2 &  Questions about world history \\
 & & {\it What were the causes of World War II? } \\
human behavior & 5 & Questions involving human emotions \\
 & & {\it I feel sad. What could I do to cheer myself up? } \\
hypothetical & 29 & Questions about hypothetical and/or counterfactual situations \\
 & & {\it If plastic was a conductor, then would a plastic spoon conduct electricity? } \\
math & 2 &  Numeric computations \\
 & & {\it What is 241 + 7864? } \\
meta-reasoning & 6 &  Questions requiring reflection about reasoning itself \\
 & & {\it What is an incorrect implication of a cat being an animal? } \\
riddle & 2 &  Trick stories with a non-obvious explanation \\
 & & {\it A young boy was rushed to the hospital emergency room, but the ER doctor saw the boy and refused to operate. "This boy is my son," the doctor said. But the doctor wasn't the boy's father. How could this be? } \\
science & 41 &  Questions in the general area of science \\
 & & {\it What gases are involved in photosynthesis? } \\
spatial & 11 &  Various spatial reasoning tasks \\
 & & {\it John is left of Sue. Where is Sue relative to John? } \\
steps & 15 & List the sequence of actions to achieve a goal \\
 & & {\it What are the steps involved in replacing a light bulb? } \\
story understanding & 25 & Tests for facts implicit in a short story \\
 & & {\it I crashed my car. When I finally left the hospital, all I wanted to do was sleep. I had to call a taxi. Why was I in hospital? } \\
temporal & 2 & Reasoning including temporal constraints (example below from \cite{davis-trials}) \\
 & & {\it Moshe posted on Facebook a photograph showing Maurice Ravel, Francois Poulenc, Frederic Mompou, and Erik Satie. Satie died in 1925. Poulenc was born in 1899. So the photograph must  have been taken when? } \\
Winograd & 3 & Winograd schema questions (requires commonsense for pronoun resolution) \cite{Levesque2011TheWS} \\
 & & {\it The elephant couldn't fit into the box because it was too big. What was too big? } \\ \hline
\end{tabular} 
} 
\caption{Categories of questions in the Challenge300 dataset. \label{challenge300}}
\end{table*}

\eat{
\begin{table}
\centering
{\small
\setlength{\tabcolsep}{3pt}	
\begin{tabular}{lcccccc}
\hline
{\bf Angle$\rightarrow$} & QC & QC &  QEC & Q & Q &  QE\\
{\bf Model$\downarrow$} & $\rightarrow${\bf A}E & 
$\rightarrow${\bf A} &  $\rightarrow${\bf A} &
$\rightarrow${\bf A}E & 
$\rightarrow${\bf A} &  $\rightarrow${\bf A}\\
\hline
\macaw{}  (11B) & 81.1 & 82.3 & 87.3 & 77.1 & 78.1 & 85.2\\
\macaw{}-3B & tbd & tbd & tbd & tbd & tbd & tbd \\
\macaw{}-large & tbd & tbd & tbd & tbd & tbd & tbd \\
\hline
\end{tabular}
}
\caption{Performance of the answer output slot on ARC-DA development set, using manual evaluation of answer quality. \oyvind{The tbd's here require some further manual evaluation by me on the smaller models - will do that if we keep this table}
\label{arc-da-eval}}
\end{table}
}

\eat{
\begin{table}
\centering
{\small
\setlength{\tabcolsep}{3pt}	
\begin{tabular}{lcccccc}
\hline
{\bf Angle$\rightarrow$} & QMC & AC &  AC & - &  - &  -\\
{\bf Model$\downarrow$} & $\rightarrow$A{\bf E} & 
$\rightarrow${\bf Q}M &  $\rightarrow$Q{\bf M} &
- & 
- & -\\
\hline
\macaw{}  (11B) & 27.6 & 48.8 & tbd & - & - & -\\
\macaw{}-3B & 28.4 & 48.0 & 68.0 & - & - & - \\
\macaw{}-large & tbd & tbd & tbd & - & - & - \\
\hline
\end{tabular}
}
\caption{Evaluation of non-answer output slots for selected angles on ARC  MC development set. \oyvind{TODO maybe? The numbers are weird though, and not even trending worse for 3B vs 11B model}
\label{arc-mc-eval-other-angles}}
\end{table}
}

\begin{table}
\centering
{\small
\begin{tabular}{lcc}
\hline
{\bf Model} & Score (\%) & \# incoherent \\ \hline
T5-CBQA (T5.1.1.XXL, NaturalQ) & 57.3 & 28 \\
Jurassic-1 (jumbo, T=0)     & 64.9 & 12 \\
GPT-3 (davinci T=0)         & 64.9 & 10 \\
\macaw{} (11B)              & 75.0 & 2 \\
\hline
\end{tabular}
}
\caption{Scores on the Challenge300 dataset, plus absolute
number of incoherent (nonsensical) answers produced. \macaw{} significantly
outperforms the other systems on this dataset.
All models are applied zero-shot. 
\label{challenge300-results}}
\end{table}

\subsection{The Challenge300 Dataset \label{challenge300-section}}

We also assembled a dataset of 300 challenge questions, called Challenge300, based on our attempts to
``break'' \macaw{} using a wide variety of question types. Most of the questions were created from scratch,
probing different styles of problem, plus a handful were drawn from the excellent challenge questions in
\cite{Davis2016HowTW} and \cite{davis-trials}.
We recorded all the questions tried (both those \macaw{} got right, and those it got wrong), rather than
cherry-picking good/bad cases. We also performed a loose classification of those questions into 22
different categories, described in Table~\ref{challenge300}. Note that this categorization is somewhat
approximate, as questions can fall into more than one category (in such cases, we attempted to
select the dominant category). However, it is still informative for analyzing the successes
and failures of \macaw{}, which we discuss in detail in Section~\ref{analysis} shortly.

For comparison, we also gave the Challenge300 questions to
T5-CBQA (size XXL)\footnote{The most powerful CBQA model, built on T5-11B with further pretraining using salient span masking (SSM), https://huggingface.co/google/t5-xxl-ssm-nq},
GPT-3 (davinci)\footnote{With prompt simply "Q: $<${\it question}$>$ A:". The GPT-3 continuation invariably contains the answer followed by more QA pairs labeled "Q:" "A:". We truncate off those additional QA pairs.},
and the recent Jurassic-1 (jumbo) model from AI21 Labs\footnote{https://www.ai21.com/blog/announcing-ai21-studio-and-jurassic-1}.
As the questions are direct answer (``open response''), with (typically) multiple, valid answers,
we scored the answers manually.
We also noted the (rare) cases where an answer was incoherent, i.e., was nonsensical (rather than just wrong). For example,
for the question "Mary owns a canary named Paul. Did Paul has any ancestors who were alive in the year 1750?" the answer "Paul" 
was marked as incoherent.
The Challenge300 dataset, answers given by each of the four systems tested, and their scores are available as part of the \macaw{} release.

The results on this dataset are shown in Table~\ref{challenge300-results}. 
On this small but challenging test, \macaw{} significantly outperforms the other systems (+10\% absolute),
as well as having fewer incoherent answers.

We also computed the average scores per question category, shown in Figure~\ref{challenge300-results-figure}, ignoring categories with fewer than five
questions (where averages are more unreliable).
The tabular version of this data, including all categories, is also provided in the Appendix.
At a high level, the general trends are similar among the different language models tested, e.g., they all tend to
excel at general knowledge questions, but struggle with questions based on false presuppositions.

\section{Qualitative Results and Analysis \label{analysis}}

We now present a qualitative analysis of \macaw{}'s strengths and weaknesses,
based on the Challenge300 questions. Note that most of these questions are
well outside the domain of the training questions, and no other context
is supplied to \macaw{} to answer them beyond the questions themselves.
Also note that all these examples are using \macaw{} zero-shot.

We first present some of the impressive (to us) examples of \macaw{}'s behavior. 
We then show some cases where it produces wrong answers, and characterize those failures.

\subsection{\macaw{}: Good Examples \label{impressive}}

\noindent
Below, Q denotes the question posed to \macaw{}, while A denotes \macaw{}'s generated answer.
We also tentatively speculate about what skills the answers {\it appear} to reveal.

\vspace{1mm}

\noindent
    {\bf 1. Entity Substitution:} Identifying a suitable replacement item. We observe similar good answers for
    several other probes (e.g., if no stove, use a microwave).

\quotebox{
Q: James went camping in the woods, but forgot to bring a hammer to bang the tent pegs in. What else might he use? \\
$\rightarrow$ A: rocks \\
Q: James went camping in the woods, but forgot to bring a hammer to bang the tent pegs in. Why might a rock be a useful substitute for the hammer? \\
$\rightarrow$ A: it is hard }

\noindent {\bf 2. Hypotheticals:} \macaw{} shows some ability to reason about novel (hypothetical/counterfactual) situations.
The second constrast query below suggests that \macaw{} was doing more than just guessing.

\quotebox{
Q: If plastic was a conductor, then would a plastic spoon conduct electricity? \\
$\rightarrow$ A: yes \\
Q: If plastic was an insulator, then would a plastic spoon conduct electricity? \\
$\rightarrow$ A: no}

Similarly, \macaw{} gives a suprisingly insightful answer about a novel situation:

\quotebox{
Q: How would you make a house conduct electricity? \\
$\rightarrow$ A: paint it with a metal paint}

Finally the two examples below hints at some awareness of the role of an object (here, a roof) in a hypothetical situation:

\quotebox{
Q: If my house did not have a roof, and it rained, what would happen to my bed upstairs?\\
$\rightarrow$ A: it would get wet\\
Q: If my house had a roof, and it rained, what would happen to my bed upstairs? \\
$\rightarrow$ A: it would stay dry}

as well as correctly reasoning about a hypothetical bird:

\quotebox{
Q: If a bird didn't have wings, how would it be affected? \\
$\rightarrow$ A: it would be unable to fly}

\paragraph {\bf 3. Meta-reasoning:} This surprising behavior hints at knowing what an implication is, and being able
to illustrate that knowledge in a specific context.

\quotebox{
Q: What is an implication of a cat being an animal? \\
$\rightarrow$ A: it has feelings \\
Q: What is an incorrect implication of a cat being an animal? \\
$\rightarrow$ A: it is a human}

\paragraph {\bf 4. False presuppositions:} Questions based on a false premise, such as the below, frequently
fool language models \cite{Kim2021WhichLI}. The example below is somewhat well-known, e.g., Google
(incorrectly) answers 1995, but \macaw{} surprisingly handles this gracefully. In general, however,
\macaw{} is fooled by these kind of examples (Section~\ref{unimpressive}).

\quotebox{
Q: What year did Tom Hanks land on the moon? \\
$\rightarrow$ A: no year}

\paragraph{\bf 5. Story Understanding:}

In our tests, \macaw{} generally answers reading comprehension questions about short stories well, for example:

\quotebox{
  Q: Karen was assigned a roommate her first year of college. Her roommate asked her to go to a nearby city for a concert. Karen agreed happily.  The show was absolutely exhilarating. What happened that might cause Karen to become good friends with her roommate? \\
$\rightarrow$ A: they had a good time together}

\paragraph {\bf 6. Commonsense:} The below suggests \macaw{} has some
commonsense awareness about the relationships between melting, temperature, and fridges.

\quotebox{
Q: If I put some cheese in the fridge, will it melt?\\
$\rightarrow$ A: no, because it will be too cold}

\paragraph {\bf 7. Comparisions:} \macaw{} is able to answer unusual comparison questions, such as below.
Such questions are interesting as they are unlikely to have been seen in \macaw{}'s/T5's training corpora.

\quotebox{
Q: How do pandas and parrots differ? \\
$\rightarrow$ A: pandas have black and white fur, while parrots have brightly colored feathers and are usually green or yellow in color}

\subsection{\macaw{}: Example Failures \label{unimpressive}}

\macaw{} answered about 24\% of the Challenge300 questions incorrectly. Below we show some examples of these
failures, and categorize the nature of the problem that it struggles with.
Again, note that all these examples are using \macaw{} zero-shot.

\vspace{1mm}

\noindent {\bf 1. Non-trivial Arithmetic:} \macaw{} generally fails at non-trivial arithmetic, although it appears to 
know the answer is some number. \macaw{} {\it will} handle simple arithmetic correctly,
e.g., Q: What is 7 + 5? $rightarrow$ 12 [correct];
Q: What is 25 + 59? $\rightarrow$ 84 [correct], but with larger numbers (3 or more digits) it typically fails. Interestingly, the answers are often in the
right ballpark, e.g., below. Similarly, Q: What is 42623 + 543456? $\rightarrow$ 870315 [incorrect, though the right
  order of magnitude].
\quotebox{
Q: What is 241 + 7864? \\
$\rightarrow$ A: 12345  \red{[wrong]}}

\noindent {\bf 2. Entity Tracking and State Changes:}
\macaw{} struggles with stories about a changing world state.
The below example is similar to those in the bAbI dataset (Tasks 2 and 3) \cite{bAbi}.
While models can be trained for reasoning about processes and a changing world state,
e.g., \cite{Henaff2017TrackingTW,Dalvi2018TrackingSC,Tandon2018ReasoningAA},
\macaw{} does not have this capability zero-shot.

\quotebox{
Q: John was in the bedroom. John put a coin in a bag. Then John took the bag to the kitchen. Then John took the coin out of the bag. John took the bag to the garden. Where was the bag at the end? \\
$\rightarrow$ A: in the kitchen	\red{[wrong]}}

Similarly, \macaw{} struggles to combine counting and object tracking, as illustrated below.

\quotebox{
Q: Imagine an empty cup. Now put a coin in the cup. Now take the coin out of the cup. How many coins are in the cup now? \\
$\rightarrow$ A: 1 \red{[wrong]}}

A final example of struggling to track worlds is below:

\quotebox{
Q: My house is red. I painted my house white. What color is my house now? \\
$\rightarrow$ A: it is now white \green{[correct]} \\
Q: My house is blue. I painted my house red. What color was my house before I painted it? \\
$\rightarrow$ A: blue \green{[correct]} \\
Q: My house is red. My neighbor's house is blue. I painted my house white. What color is my house now? \\
$\rightarrow$ A: it's the same color as my neighbor's house \red{[wrong]} \\
Q: My house is red. I painted my neighbor's house white. What color is my house now? \\
$\rightarrow$ A: half red, half white  \red{[wrong]}}

\noindent {\bf 3. Spatial Reasoning:}

\macaw{} is unreliable at non-trivial spatial reasoning, as illustrated below, suggesting
it is not constructing a latent spatial layout during question interpretation:

\quotebox{
Q: John is left of Sue. Where is Sue relative to John? \\
$\rightarrow$ A: to his right \green{[correct]} \\ 
Q: John is left of Sue. Sue is left of Mike. Where is Mike relative to John? \\
$\rightarrow$  A: to his right \green{[correct]} \\
Q: John is left of Sue. Sue is left of Mike. Where is Sue relative to John? \\
$\rightarrow$  A: to his left	\red{[wrong]}}

\noindent {\bf 4. Hypotheticals: Meaning-Shifting Modifiers} Below, a subtle change (``{\it electric} car'') fools \macaw{}.

\quotebox{
Q: If an electric car did not have any gas in it, how would it be affected? \\
$\rightarrow$ A: it would not move  \red{[wrong]}}

\noindent {\bf 5. Some Winograd Schema:} Winograd Schema questions ask a model to correctly resolve a pronoun,
typically requiring commonsense \cite{Levesque2011TheWS}.
While \macaw{} generally answers them correctly, it can make mistakes such as the example below.

\quotebox{
Q: The elephant couldn't fit into the box because it was too big. What was too big? \\
$\rightarrow$ A: the box 	 \red{[wrong]}}

\noindent {\bf 6. False Presuppositions:} In general, \macaw{} is fooled by questions with false presuppositions,
such as below. (A counterexample was shown earlier in Section~\ref{impressive} where \macaw{} noted Tom Hanks has not been to the moon).

\quotebox{
Q: How old was Mark Zuckerberg when he founded Google? \\
-$\rightarrow$ A: 17   \red{[wrong, trick question]}}

\noindent {\bf 7. Commonsense:} While \macaw{} correctly answers many questions apparently requiring commonsense, it
can make errors, such as the below:

\quotebox{
Q: What happens if I drop a  glass on a bed of feathers? \\
$\rightarrow$ A: the glass shatters \red{[wrong]}}

Similarly, \macaw{} fails the following false presupposition, commonsense question adapted from from \cite{Davis2016HowTW}:

\quotebox{
Q: Sally's favorite cow died yesterday. When will the cow be alive again? \\
$\rightarrow$ A: in a few days \red{[wrong]}}

Marcus and Davis have similarly probed GPT3 with other commonsense problems (asking for a sensible continuation
of a setup), and note that GPT3 fails a substantial number of them \cite{bloviator}.
While we have not probed such commonsense capabilities systematically for \macaw{}, 
we expect it will similarly fail many of them also. We note that in our results by category
in Figure~\ref{challenge300-results}, all the models perform relatively poorly on commonsense questions (compared
with other types).

Orthogonal to our question categories, we also note the following interesting behaviors of \macaw{}:

\noindent {\bf Overly Brief Answers:} \macaw{} typically generates short answers, sometimes so short that they can
barely be considered correct, such as below:

\quotebox{
Q: How does photosynthesis work? \\
$\rightarrow$ A: sun shines on plants \red{[highly incomplete]}}

\noindent {\bf Iterative queries:} Below, we ask a simple question, then re-ask the question but ask for a different answer.
\macaw{} is able to sustain this for several iterations, but eventually breaks down. (Below, the manual question and \macaw{} answer
are shown together on the same line.)

\quotebox{
What is made of metal? a car \\
What is made of metal, besides a car? a door \\
What is made of metal, besides a car and a door? a bicycle \\
What is made of metal, besides a car, a door, and a bicycle? a spoon \\
What is made of metal, besides a car, a door, a bicycle, and a spoon? a spoon \red{[Now repeating an answer]}}

\noindent {\bf Generating Narratives:} We can similarly ask \macaw{} to generate a plausible event sequence (``story'') by iteratively giving a scenario, asking ``What happens next?'', and then adding the answer back into the question and re-asking it. For example:

\quotebox{
Some kids are planning a rollerskating race. What happens next? They practice. \\
Some kids are planning a rollerskating race. They practice. What happens next? They fall. \\
Some kids are planning a rollerskating race. They practice. They fall. What happens next? ...}

Eventually \macaw{} starts repeating answers, as illustrated below as a continuation of the earlier questions.
The sequence of events in the question below reflect \macaw{}'s earlier answers to the ``....What happens next?'' questions.

\quotebox{
  Some kids are planning a rollerskating race. They practice. They fall. They get up and try again. They fall again. They give up. They lose interest in the sport. They stop trying. They never learn. They never learn. They never learn. They never learn. ... What happens next? They never learn.}

While a possibly plausible sequence, this is hardly a good story.

\subsection{Other Models' Answers}

While our focus here is on \macaw{}, we note that the three other models tested (GPT-3, T5-CBQA, and Jurassic-1) similarly
exhibit moments of both brilliance and ignorance on Challenge300, with overall lower scores than \macaw{} (Table~\ref{challenge300-results}).
The full list of all the models' answers is included in the \macaw{} release.

\subsection{Harmful and Offensive Answers: A Note of Caution}

As a final word of caution: like other pretrained models that have seen potentially
harmful/offensive text in pretraining, \macaw{} is capable of producing biased
and/or offensive answers depending on the question, a phenomenon of concern
and significant attention in the community, e.g.,
\cite{Li2020UnQoveringSB,Zhou2021ChallengesIA}. Care must be used when
deploying large-scale language models such as \macaw{} in practical settings.

\section{Summary}

To assist other researchers, we have released \macaw{}, a high-quality, T5-based QA system
that exemplifies both the power and limits of current pretrained language models.
  \macaw{} exhibits strong performance, zero-shot, on a wide variety of topics, 
  including outperforming GPT-3 by over 10\% (absolute) on Challenge300, a suite of 300 challenge questions,
  despite being an order of magnitude smaller (11 billion vs. 175 billion parameters).
  In addition, a \macaw{}-based model currently tops the leaderboards on the ARC datasets (Section~\ref{arc}).
    One might consider \macaw{} as a language model highly optimized for question-answering tasks,
    including allowing different permutations of input/output slots (``angles'') related to
    question-answering.
    
We have also illustrated some surprisingly impressive answers \macaw{} produces,
as well as some categories of questions that it still struggles with,
providing insights into the strengths and weaknesses of \macaw{} and likely other transformer-based QA systems.
We hope that \macaw{} proves useful to the community, both as a zero-shot QA system,
and as a strong starting point for further fine-tuning on specific tasks where
training data is available and the highest precision possible is required.
\macaw{} is available at~\macawurl.


\eat{
Modern language models are generally good at question-answering, and \macaw{} embodies much of that capability
in a well-tuned, general-purpose, freely available system. We have illustrated some surprisingly impressive
answers it produces, causing us to ask the question: to what extent (if any) is \macaw{} constructing a ``mental
model'' of the world inside itself? From answers such as those in Figure~\ref{examples}, it is hard to
imagine that there isn't {\it some} sort of neural reasoning or modeling going on internally,
even if vague and incomplete. On the other hand, as Section~\ref{failures} shows, the model can also
go substantially wrong on occasions. We hope further probing helps answer such questions.
\macaw{} is available at~\macawurl.
}



\bibliography{references}

\begin{thebibliography}{26}
\expandafter\ifx\csname natexlab\endcsname\relax\def\natexlab#1{#1}\fi
\expandafter\ifx\csname url\endcsname\relax
  \def\url#1{{\tt #1}}\fi

\bibitem[Bhakthavatsalam et~al.(2021)Bhakthavatsalam, Khashabi, Khot, Mishra,
  Richardson, Sabharwal, Schoenick, Tafjord, and Clark]{arc-da}
S.~Bhakthavatsalam, D.~Khashabi, T.~Khot, B.~D. Mishra, K.~Richardson,
  A.~Sabharwal, C.~Schoenick, O.~Tafjord, and P.~Clark.
\newblock Think you have solved direct-answer question answering? try arc-da,
  the direct-answer ai2 reasoning challenge.
\newblock {\em ArXiv}, abs/2102.03315, 2021.

\bibitem[Brown et~al.(2020)Brown, Mann, Ryder, Subbiah, Kaplan, Dhariwal,
  Neelakantan, Shyam, Sastry, Askell, Agarwal, Herbert-Voss, Krueger, Henighan,
  Child, Ramesh, Ziegler, Wu, Winter, Hesse, Chen, Sigler, Litwin, Gray, Chess,
  Clark, Berner, McCandlish, Radford, Sutskever, and Amodei]{gpt3}
T.~Brown, B.~Mann, N.~Ryder, M.~Subbiah, J.~Kaplan, P.~Dhariwal,
  A.~Neelakantan, P.~Shyam, G.~Sastry, A.~Askell, S.~Agarwal, A.~Herbert-Voss,
  G.~Krueger, T.~Henighan, R.~Child, A.~Ramesh, D.~M. Ziegler, J.~Wu,
  C.~Winter, C.~Hesse, M.~Chen, E.~Sigler, M.~Litwin, S.~Gray, B.~Chess,
  J.~Clark, C.~Berner, S.~McCandlish, A.~Radford, I.~Sutskever, and D.~Amodei.
\newblock Language models are few-shot learners.
\newblock In {\em NeurIPS}, 2020.

\bibitem[Clark et~al.(2018)Clark, Cowhey, Etzioni, Khot, Sabharwal, Schoenick,
  and Tafjord]{Clark2018ThinkYH}
P.~Clark, I.~Cowhey, O.~Etzioni, T.~Khot, A.~Sabharwal, C.~Schoenick, and
  O.~Tafjord.
\newblock Think you have solved question answering? {T}ry {ARC}, the {AI2}
  {R}easoning {C}hallenge.
\newblock {\em ArXiv}, abs/1803.05457, 2018.

\bibitem[Clark et~al.(2016)Clark, Etzioni, Khot, Sabharwal, Tafjord, Turney,
  and Khashabi]{Clark2016CombiningRS}
P.~Clark, O.~Etzioni, T.~Khot, A.~Sabharwal, O.~Tafjord, P.~D. Turney, and
  D.~Khashabi.
\newblock Combining retrieval, statistics, and inference to answer elementary
  science questions.
\newblock In {\em AAAI}, 2016.

\bibitem[Dalvi et~al.(2018)Dalvi, Huang, Tandon, tau Yih, and
  Clark]{Dalvi2018TrackingSC}
B.~Dalvi, L.~Huang, N.~Tandon, W.~tau Yih, and P.~Clark.
\newblock Tracking state changes in procedural text: a challenge dataset and
  models for process paragraph comprehension.
\newblock In {\em NAACL-HLT}, 2018.

\bibitem[Davis(2016)]{Davis2016HowTW}
E.~Davis.
\newblock How to write science questions that are easy for people and hard for
  computers.
\newblock {\em AI Mag.}, 37:\penalty0 13--22, 2016.

\bibitem[Hase et~al.(2020)Hase, Zhang, Xie, and
  Bansal]{Hase2020LeakageAdjustedSC}
P.~Hase, S.~Zhang, H.~Xie, and M.~Bansal.
\newblock Leakage-adjusted simulatability: Can models generate non-trivial
  explanations of their behavior in natural language?
\newblock In {\em EMNLP}, 2020.

\bibitem[Henaff et~al.(2016)Henaff, Weston, Szlam, Bordes, and
  LeCun]{Henaff2017TrackingTW}
M.~Henaff, J.~Weston, A.~D. Szlam, A.~Bordes, and Y.~LeCun.
\newblock Tracking the world state with recurrent entity networks.
\newblock In {\em ICLR}, 2016.

\bibitem[Holtzman et~al.(2020)Holtzman, Buys, Forbes, and
  Choi]{Holtzman2020TheCC}
A.~Holtzman, J.~Buys, M.~Forbes, and Y.~Choi.
\newblock The curious case of neural text degeneration.
\newblock {\em ArXiv}, abs/1904.09751, 2020.

\bibitem[Jansen et~al.(2018)Jansen, Wainwright, Marmorstein, and
  Morrison]{Jansen2018WorldTreeAC}
P.~A. Jansen, E.~Wainwright, S.~Marmorstein, and C.~T. Morrison.
\newblock Worldtree: A corpus of explanation graphs for elementary science
  questions supporting multi-hop inference.
\newblock In {\em LREC}, 2018.
\newblock Also arXiv:1802.03052.

\bibitem[Kalyan et~al.(2021)Kalyan, Kumar, Chandrasekaran, Sabharwal, and
  Clark]{fermi}
A.~Kalyan, A.~Kumar, A.~Chandrasekaran, A.~Sabharwal, and P.~Clark.
\newblock How much coffee was consumed during emnlp 2019? fermi problems: A new
  reasoning challenge for ai.
\newblock In {\em EMNLP}, 2021.

\bibitem[Khashabi et~al.(2020{\natexlab{a}})Khashabi, Min, Khot, Sabharwal,
  Tafjord, Clark, and Hajishirzi]{unifiedqa}
D.~Khashabi, S.~Min, T.~Khot, A.~Sabharwal, O.~Tafjord, P.~Clark, and
  H.~Hajishirzi.
\newblock Unifiedqa: Crossing format boundaries with a single qa system.
\newblock In {\em EMNLP}, 2020{\natexlab{a}}.

\bibitem[Khashabi et~al.(2020{\natexlab{b}})Khashabi, Min, Khot, Sabharwal,
  Tafjord, Clark, and Hajishirzi]{Khashabi2020UnifiedQACF}
D.~Khashabi, S.~Min, T.~Khot, A.~Sabharwal, O.~Tafjord, P.~Clark, and
  H.~Hajishirzi.
\newblock Unifiedqa: Crossing format boundaries with a single {QA} system.
\newblock In {\em EMNLP}, 2020{\natexlab{b}}.

\bibitem[Kim et~al.(2021)Kim, Pavlick, Ayan, and Ramachandran]{Kim2021WhichLI}
N.~Kim, E.~Pavlick, B.~K. Ayan, and D.~Ramachandran.
\newblock Which linguist invented the lightbulb? presupposition verification
  for question-answering.
\newblock {\em ArXiv}, abs/2101.00391, 2021.

\bibitem[Kwiatkowski et~al.(2019)Kwiatkowski, Palomaki, Redfield, Collins,
  Parikh, Alberti, Epstein, Polosukhin, Kelcey, Devlin, Lee, Toutanova, Jones,
  Chang, Dai, Uszkoreit, Le, and Petrov]{47761}
T.~Kwiatkowski, J.~Palomaki, O.~Redfield, M.~Collins, A.~Parikh, C.~Alberti,
  D.~Epstein, I.~Polosukhin, M.~Kelcey, J.~Devlin, K.~Lee, K.~N. Toutanova,
  L.~Jones, M.-W. Chang, A.~Dai, J.~Uszkoreit, Q.~Le, and S.~Petrov.
\newblock Natural questions: a benchmark for question answering research.
\newblock {\em Transactions of the Association of Computational Linguistics},
  2019.

\bibitem[Lai et~al.(2017)Lai, Xie, Liu, Yang, and Hovy]{race}
G.~Lai, Q.~Xie, H.~Liu, Y.~Yang, and E.~Hovy.
\newblock {RACE}: {L}arge-scale reading comprehension dataset from
  examinations.
\newblock In {\em EMNLP}, 2017.

\bibitem[Levesque et~al.(2011)Levesque, Davis, and
  Morgenstern]{Levesque2011TheWS}
H.~Levesque, E.~Davis, and L.~Morgenstern.
\newblock The winograd schema challenge.
\newblock In {\em KR}, 2011.

\bibitem[Li et~al.(2020)Li, Khashabi, Khot, Sabharwal, and
  Srikumar]{Li2020UnQoveringSB}
T.~Li, D.~Khashabi, T.~Khot, A.~Sabharwal, and V.~Srikumar.
\newblock Unqovering stereotyping biases via underspecified questions.
\newblock In {\em EMNLP}, 2020.

\bibitem[Marcus and Davis(2020{\natexlab{a}})]{davis-trials}
G.~Marcus and E.~Davis.
\newblock Experiments testing gpt-3's ability at commonsense reasoning:
  Results.
\newblock Technical report, NYU, 2020{\natexlab{a}}.
\newblock (https://cs.nyu.edu/$\sim$davise/papers/GPT3CompleteTests.html).

\bibitem[Marcus and Davis(2020{\natexlab{b}})]{bloviator}
G.~Marcus and E.~Davis.
\newblock Gpt-3, bloviator: Openai’s language generator has no idea what
  it’s talking about.
\newblock {\em MIT Technology Review}, Aug 2020{\natexlab{b}}.

\bibitem[Radford et~al.(2018)Radford, Narasimhan, Salimans, and
  Sutskever]{gpt2}
A.~Radford, K.~Narasimhan, T.~Salimans, and I.~Sutskever.
\newblock Improving language understanding by generative pre-training.
\newblock Technical report, OpenAI, 2018.

\bibitem[Raffel et~al.(2020)Raffel, Shazeer, Roberts, Lee, Narang, Matena,
  Zhou, Li, and Liu]{Raffel2020ExploringTL}
C.~Raffel, N.~Shazeer, A.~Roberts, K.~Lee, S.~Narang, M.~Matena, Y.~Zhou,
  W.~Li, and P.~J. Liu.
\newblock Exploring the limits of transfer learning with a unified text-to-text
  transformer.
\newblock {\em J. Mach. Learn. Res.}, 21:\penalty0 140:1--140:67, 2020.

\bibitem[Roberts et~al.(2020)Roberts, Raffel, and Shazeer]{Roberts2020HowMK}
A.~Roberts, C.~Raffel, and N.~M. Shazeer.
\newblock How much knowledge can you pack into the parameters of a language
  model?
\newblock In {\em EMNLP}, 2020.

\bibitem[Tandon et~al.(2018)Tandon, Dalvi, Grus, tau Yih, Bosselut, and
  Clark]{Tandon2018ReasoningAA}
N.~Tandon, B.~Dalvi, J.~Grus, W.~tau Yih, A.~Bosselut, and P.~Clark.
\newblock Reasoning about actions and state changes by injecting commonsense
  knowledge.
\newblock In {\em EMNLP}, 2018.

\bibitem[Weston et~al.(2016)Weston, Bordes, Chopra, and Mikolov]{bAbi}
J.~Weston, A.~Bordes, S.~Chopra, and T.~Mikolov.
\newblock Towards {AI-Complete} question answering: A set of prerequisite toy
  tasks.
\newblock In {\em ICLR}, 2016.

\bibitem[Zhou et~al.(2021)Zhou, Sap, Swayamdipta, Smith, and
  Choi]{Zhou2021ChallengesIA}
X.~Zhou, M.~Sap, S.~Swayamdipta, N.~A. Smith, and Y.~Choi.
\newblock Challenges in automated debiasing for toxic language detection.
\newblock In {\em EACL}, 2021.

\end{thebibliography}
\bibliographystyle{myabbrvnat}

\appendix

\onecolumn

\section*{Appendix: Average Scores of Models on the Challenge300 Question Categories}

\vspace{1cm}

Table~\ref{numbers} provides the histogram data from Figure~\ref{challenge300-results-figure} in tabular form, plus
remaining question categories with fewer than 5 questions that were not included in the histogram (where average scores may be unreliable).

\begin{table*}[h]
\centering
{\small
  \begin{tabular}{ll|llll|l} \hline
       & & \multicolumn{4}{|c|}{\bf Model} & {\bf Average of} \\ 
  {\bf Qn Category} & {\bf \# Qns} & {\bf \macaw{}} & {\bf GPT-3} & {\bf T5-CBQA} & {\bf Jurassic-1} & {\bf Averages} \\ \hline
commonsense	& 38	& 0.50	& 0.53	& 0.42	& 0.47	& 0.48 \\
comparison	& 2	& 1.00	& 0.50	& 0.50	& 1.00	& 0.75 \\
entity substitution	& 4	& 1.00	& 0.63	& 1.00	& 1.00	& 0.91 \\
entity tracking	& 13	& 0.50	& 0.62	& 0.65	& 0.54	& 0.58 \\
estimation	& 4	& 0.88	& 1.00	& 0.75	& 0.50	& 0.78 \\
example generation	& 2	& 1.00	& 1.00	& 0.50	& 0.00	& 0.63 \\
explanation	& 14	& 0.68	& 0.43	& 0.68	& 0.64	& 0.61 \\
false presupposition	& 9	& 0.11	& 0.00	& 0.00	& 0.00	& 0.03 \\
general knowledge	& 70	& 0.93	& 0.79	& 0.73	& 0.80	& 0.81 \\
generation	& 1	& 1.00	& 1.00	& 0.00	& 1.00	& 0.75 \\
history	& 2	& 1.00	& 1.00	& 1.00	& 1.00	& 1.00 \\
human behavior	& 5	& 0.70	& 0.60	& 0.90	& 0.60	& 0.70 \\
hypothetical	& 29	& 0.78	& 0.59	& 0.71	& 0.52	& 0.65 \\
math	& 2	& 0.00	& 0.50	& 0.00	& 0.00	& 0.13 \\
meta-reasoning	& 6	& 1.00	& 0.67	& 0.67	& 0.33	& 0.67 \\
riddle	& 2	& 1.00	& 0.50	& 0.00	& 0.50	& 0.50 \\
science	& 41	& 0.76	& 0.65	& 0.43	& 0.74	& 0.65 \\
spatial	& 11	& 0.73	& 0.77	& 0.45	& 0.82	& 0.69 \\
steps	& 15	& 0.87	& 0.73	& 0.31	& 0.77	& 0.67 \\
story understanding	& 25	& 0.88	& 0.72	& 0.77	& 0.72	& 0.77 \\
temporal	& 2	& 0.25	& 0.00	& 0.25	& 0.25	& 0.19 \\
Winograd	& 3	& 0.67	& 1.00	& 0.00	& 1.00	& 0.67  \\ \hline 
{\bf ALL}	& 300	& 0.75	& 0.65	& 0.57	& 0.65  & 0.66 \\ \hline
\end{tabular} 
} 
\caption{Average score of models on different question categories in Challenge300. (See Figure~\ref{challenge300-results-figure} for histogram). \label{numbers}} 
\end{table*} 
 
\end{document}